\pdfoutput=1

\documentclass[11pt]{article}

\usepackage[final]{acl}

\usepackage{microtype}

\usepackage{times}
\usepackage{latexsym}
\usepackage{booktabs, tabularx}
\usepackage{makecell}
\usepackage{amsmath}
\usepackage{slashbox}
\usepackage[utf8]{inputenc}
\usepackage[arabic,farsi,russian,english]{babel}

\usepackage{microtype}

\usepackage{pgfplots}
\usepgfplotslibrary{fillbetween}
\usetikzlibrary{patterns}
\usetikzlibrary{pgfplots.groupplots}
\pgfplotsset{compat=1.17}
\usepackage{lscape}
\usepackage{multirow}
\usepackage{rotating}
\usepackage{dblfloatfix}
\usepackage{xcolor}
\usepackage{pgf}
\usepackage{collcell}
\usepackage{adjustbox}
\usepackage{multirow}
\usepackage{pdflscape} 
\usepackage{booktabs}
\usepackage{colortbl}
\usepackage{xifthen}
\newcommand*{\MinNumber}{60}
\newcommand*{\MaxNumber}{95}
\definecolor{myCellColor}{RGB}{45,162,45}
\newcommand{\ApplyGradient}[1]{
  \pgfmathsetmacro{\PercentColor}{100.0*(#1-\MinNumber)/(\MaxNumber-\MinNumber)}
  \ifthenelse{\isempty{#1}}
    {}
    {
        \pgfmathparse{#1 > 60 ? 1 : 0}
        \ifthenelse{\equal{\pgfmathresult}{1}}
        {
        \pgfmathparse{#1 > 83.58 ? 1 : 0}
        \ifthenelse{\equal{\pgfmathresult}{1}}
        {\edef\x{\noexpand\cellcolor{myCellColor!\PercentColor}}\x\textcolor{black}{#1}}
        {\edef\x{\noexpand\cellcolor{myCellColor!\PercentColor}}\x\textcolor{black}{#1}}
        }
        {\edef\x{\noexpand\cellcolor{myCellColor!\PercentColor}}\x\textcolor{black}{#1}}
    }
}
\newcolumntype{R}{>{\collectcell\ApplyGradient}{r}<{\endcollectcell}}

\newcommand\blfootnote[1]{
  \begingroup
  \renewcommand\thefootnote{}\footnote{#1}
  \addtocounter{footnote}{-1}
  \endgroup
}

\usepackage{hyphenat}
\DeclareCaptionTextFormat{new}{\nohyphens{#1}}

\title{Metaphors in Pre-Trained Language Models: \\ Probing and Generalization Across Datasets and Languages}

\author{Ehsan Aghazadeh$^{\star}$ ~ Mohsen Fayyaz$^{\star}$ ~ Yadollah Yaghoobzadeh\\
School of Electrical and Computer Engineering,\\
College of Engineering, \\University of Tehran, Tehran, Iran\\
\texttt{\{eaghazade1998, mohsen.fayyaz77, y.yaghoobzadeh\}@ut.ac.ir}
}

\date{}

\begin{document}
\maketitle
\begin{abstract}
Human languages are full of metaphorical expressions.
Metaphors help people understand the world by connecting new concepts and domains to more familiar ones. 
Large pre-trained language models (PLMs) are therefore assumed to encode metaphorical knowledge useful for NLP systems.
In this paper, we investigate this hypothesis for PLMs, by probing metaphoricity information in their encodings, and by measuring the cross-lingual and cross-dataset generalization of this information.
We present studies in multiple metaphor detection datasets and in four languages (i.e., English, Spanish, Russian, and Farsi). 
Our extensive experiments suggest that contextual representations in PLMs do encode metaphorical knowledge, and mostly in their middle layers.
The knowledge is transferable between languages and datasets, especially when the annotation is consistent across training and testing sets. 
Our findings give helpful insights for both cognitive and NLP scientists.

\blfootnote{$^\star$ Equal contribution.}
\end{abstract}

\section{Introduction}

Pre-trained language models (PLMs) \cite{peters-etal-2018-deep,devlin-etal-2019-bert}, are now used in almost all NLP applications, e.g., machine translation \cite{Li2021survey}, question answering \cite{zhang2020mrc}, dialogue systems \cite{ni2021dialogue}, and sentiment analysis \cite{minaee2020classification}. 
They have sometimes been referred to as ``foundation models'' \cite{bommasani2021fm} due to their significant impact on research and industry.

\begin{figure}[t!]
\centering
    \includegraphics[width=0.48\textwidth]{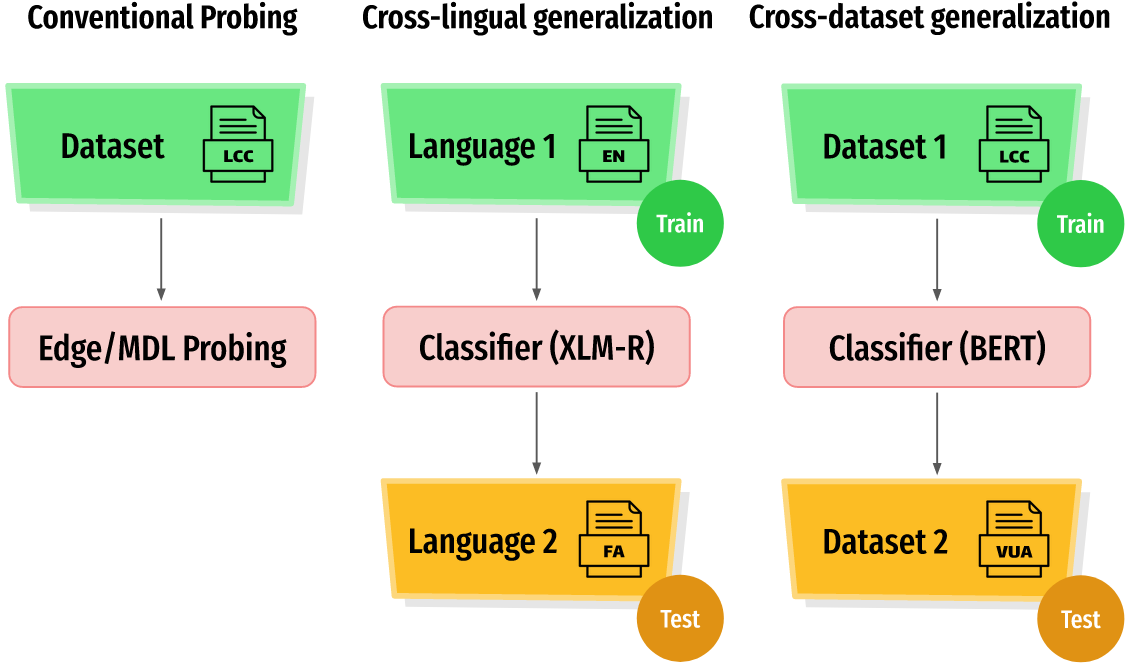}
    \caption{An illustration of our probing and generalization scenarios for metaphorical knowledge.}
    \label{fig:methods}
\end{figure}

Metaphors are important aspects of human languages.
In conceptual metaphor theory (CMT) \cite{lakoff2008metaphors}, metaphor is defined as a cognitive phenomenon associating two different concepts or domains. This phenomenon is built in cognition and expressed in language.
The creativity and problem solving (i.e., generalization to new problems) depend on the analogies and metaphors a cognitive system, like our brain, relies on.
Modeling metaphors is therefore essential in building human-like computational systems that can relate emerging concepts to the more familiar ones. 

So far, there has been no comprehensive analysis of whether and how PLMs represent metaphorical information. We intuitively assume that PLMs must encode some information about metaphors due to their great performance in metaphor detection and other language processing tasks. Confirming that experimentally is a question that we address here.
Specifically, we aim to know \emph{whether generalizable metaphorical knowledge is encoded in PLM representations or not}.
The outline of our work is presented in Figure~\ref{fig:methods}.

We first do \emph{probing} experiments to answer questions such as:
(i) with which accuracies and extractablities do different PLMs encode metaphorical knowledge? 
(ii) how deep is the metaphorical knowledge encoded in PLM multi-layer representations? 
We take two probing methods, edge probing \cite{tenney2018you} and minimum description length \cite{voita-titov-2020-information}, and apply them to four metaphor detection datasets, namely LCC \cite{mohler2016lcc}, TroFi \cite{birke-sarkar-2006-clustering}, VUA pos, and VUA Verbs \cite{steen2010method}.

To better estimate the \emph{generalization} of metaphorical knowledge in PLMs, we design two setups in which testing comes from a different distribution than training data: cross-lingual and cross-dataset metaphor detection.
Each setup can reveal important information on whether or not the metaphorical knowledge is encoded consistently in PLMs.
Four languages (English, Farsi, Russian and Spanish) and four datasets (LCC, TroFi, VUA pos, and VUA Verbs) are considered in these generalization experiments.

In summary, this paper makes the following contributions:
\begin{itemize}
    \item 
For the first time, and through careful probing analysis, we confirm that PLMs do encode metaphorical knowledge.
\item 
We show that metaphorical knowledge is encoded better in the middle layers of PLMs.
\item 
We evaluate the generalization of metaphorical knowledge in PLMs across four languages and four dataset sources, and find out that there is considerable transferability for the pairs with consistent data annotation even if they are in different languages.~\footnote{Our implementation is available at \url{https://github.com/EhsanAghazadeh/Metaphors_in_PLMs}}
\end{itemize}

\section{Related Work}

\paragraph{Metaphor detection using PLMs.}
The metaphor detection task \cite{mason-2004-cormet,birke-sarkar-2007-active,shutova-etal-2013-statistical} is a good fit for analyzing the metaphorical knowledge.
Using PLMs for metaphor detection has been common in recent years, setting new state-of-the-art results, indicating implicitly that PLMs represent metaphorical information.
\newcite{choi2021melbert} introduce a new architecture that integrates metaphor detection theories with BERT. They use the definitions as well as example usages of words jointly with PLM representations.
Similarly, \citet{song2021verb} presents a new perspective on metaphor detection task by framing it as relation classification, focusing on the verbs. 
These approaches beat the earlier work of using PLMs \cite{su-etal-2020-deepmet,chen-etal-2020-go,gong-etal-2020-illinimet}, RNN-based \cite{wu-etal-2018-neural,mao-etal-2019-end} and feature-based approaches \cite{turney-etal-2011-literal,shutova-etal-2016-black}.
Note that our goal is not to compete with these models, but to probe and analyze the relevant knowledge in PLMs.

\citet{tsvetkov14metaphor} present \textbf{cross-lingual metaphor detection} models using linguistic features and word embeddings. Bilingual dictionaries map different languages. 
Their datasets are quite small (\~1000 training and \~200 testing examples), making them unsuitable for a robust evaluation. 
However, this paper still remains as the only cross-lingual evaluation of metaphor detection, to the best of our knowledge.
Here, using multilingual PLMs, we perform zero-shot cross-lingual transfer for metaphor detection. Our goal is to test if PLMs represent metaphorical knowledge transferable across languages.

\paragraph{Probing methods in NLP.}
Probing is an analytical tool used for assessing linguistic knowledge in language representations. In probing, the information richness of the representations is inspected by the quality of a supervised model in predicting linguistic properties based only on the representations \citep{kohn-2015-whats,gupta-etal-2015-distributional,yaghoobzadeh-schutze-2016-intrinsic,conneau2018cram,tenney2018you, tenney-etal-2019-bert,yaghoobzadeh-etal-2019-probing,hewitt-manning-2019-structural,zhao-etal-2020-quantifying,belinkov2021probing}.
Here, we apply probing to perform our study on whether metaphorical knowledge is present in PLM representations, and whether that is generalizable across languages and datasets. 

\begin{figure}[t!]
\centering
    \includegraphics[width=0.48\textwidth]{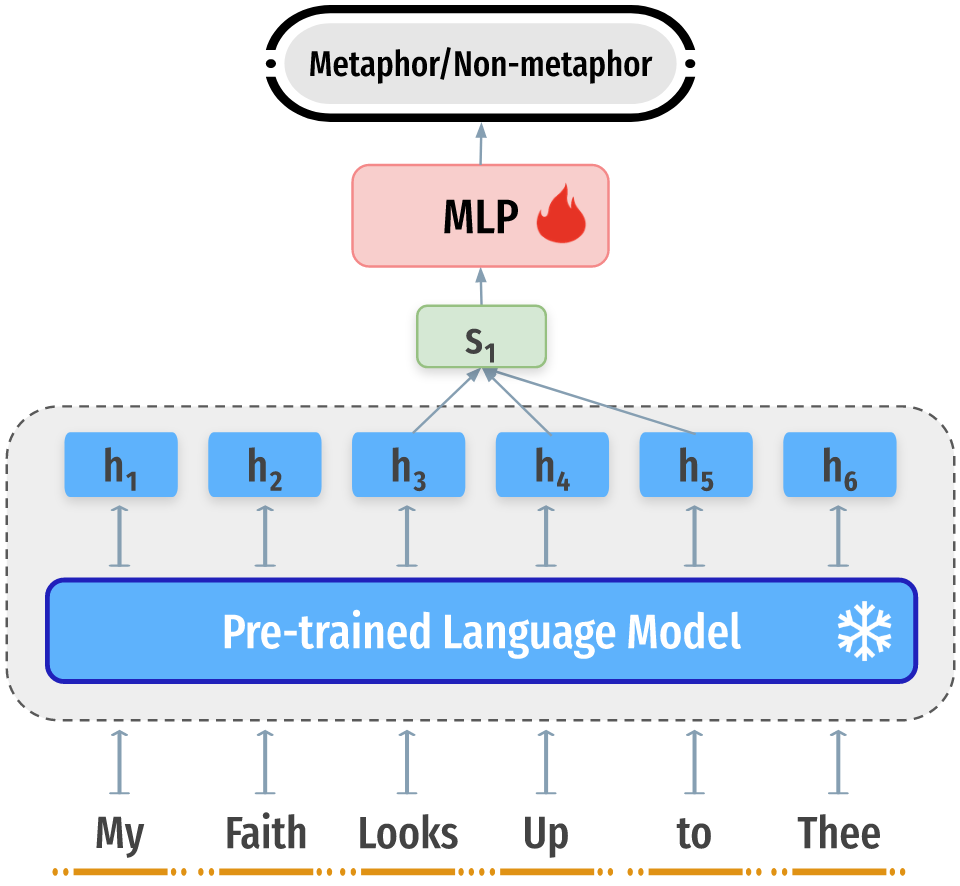}

    \caption{Probing architecture for metaphors employed in edge probing and MDL probing.}
    \label{fig:probe_architecture}
\end{figure}

A popular probing method introduced by \citet{tenney2018you} is \emph{edge probing} (Figure \ref{fig:probe_architecture}). They propose a suite of span-level tasks, including POS tagging and coreference resolution. 
Despite the widespread use of edge probing and other conventional probes, the question of whether the probing classifier is learning the task itself rather than identifying the linguistic knowledge raises concerns.

An Information-theoretic view can solve this issue \cite{voita-titov-2020-information} by reformulating probing as a data transmission problem. They consider the effort needed to extract linguistic knowledge in addition to the final quality of the probe, showing that this approach is more informative and robust than normal probing methods.
We employ both edge and MDL probing in this work.

\paragraph{Probing multilingual PLMs.}
The application of probing methods in NLP is extended to multilingual PLMs as well \cite{pires-etal-2019-multilingual,eichler-etal-2019-linspector,ravishankar-etal-2019-multilingual,ravishankar-etal-2019-probing,choenni2020what}.
\newcite{choenni2020what} introduce probing tasks for typological features of multiple languages in multilingual PLMs.
\citet{ravishankar-etal-2019-multilingual,ravishankar-etal-2019-probing} extend the probing tasks of \citet{conneau2018cram}, to a few other languages.
\citet{pires-etal-2019-multilingual} study the generalization of multilingual-BERT across
languages when performing cross-lingual downstream tasks.
Here, as part of our study, we probe the generalization of metaphorical knowledge in XLM-R \cite{conneau-etal-2020-unsupervised}, a notable multilingual PLM.

\paragraph{Out-of-distribution generalization.}
There has been no earlier work on studying or evaluating out-of-distribution generalization in metaphor detection. 
This generalization refers to scenarios where testing and training sets come from different distributions \cite{duchi2018learning,hendrycks2020many,hendrycks-etal-2020-pretrained}. 
Here, we have scenarios where testing and training data are in different languages or domains / datasets. These are challenging evaluation scenarios for the generalization of encoded information (metaphoricity in our case).

\section{Inspecting Metaphorical Knowledge in PLMs}
\label{sec:metaphor_detection}
Metaphors are used frequently in our everyday language to convey our thoughts more clearly.
There are related theories in linguistics and cognitive science. 
Following linguistic theories, metaphoricity is mostly annotated using metaphor identification procedure (MIP). MIP identifies a word in a given context as a metaphor if it has a basic or literal meaning that contrasts with its context words.
Based on conceptual metaphor theory (CMT) \cite{lakoff2008metaphors}, one target domain (e.g., ARGUMENT) is explained using a source domain (e.g., WAR). The source domain is usually more concrete or physical, while the target is more abstract.
Relating these two theories, metaphors are expressed in language connecting two contrasting domains.
For example, in ``We won the argument'', the domain of ARGUMENT is linked to the domain of WAR by using the word ``won''.
The word ``won'' is a ``metaphor'' here since its primary domain contrasts with its contextual domain.
The same word ``won'' in a sentence like ``The Allies won the war'' refers to its literal meaning and therefore is not a metaphor. 
The task of metaphor detection is defined to do this classification of ``literal'' and ``metaphor''.

Accordingly, when designing a metaphor detection system, to figure out if a token is a metaphor in a particular context, we assume following a process like:
(i) finding if the token has multiple meanings in different domains, including a more basic, concrete, or body-related meaning. For example, ``fight'', ``win'' and ``mother'' satisfy this condition.
(ii) finding if the source domain of the token contrasts with the target domain. Here the contrast is important and finding the exact domains might not be necessary. The source domain, in which its literal / basic meaning resides, is a non-contextual attribute, while the target domain is mainly found using the contextual clues (WAR and ARGUMENT for ``won'' in the above example).
  
Here, we use the metaphor detection datasets annotated based on these theories and analyze the PLM representations to see if they encode metaphorical knowledge and if the encoding is generalizable.
To do so, we first probe PLMs for their metaphorical information, generally and also across layers. This gives us intuition on how well metaphoricity is encoded and how local or contextual that is. 
Then, we test if the knowledge of metaphor detection can be transferred across languages and if multilingual PLMs capture that. 
Finally, the generalization of metaphorical knowledge across datasets is examined to see if the theories and annotations followed by different datasets are consistent, and if PLMs learn generalizable knowledge rather than dataset artifacts.

\subsection{Probing}
\label{sec:probing_intro}

Here, we aim to answer general questions about metaphors in PLMs: do PLMs encode metaphorical information and, if so, how it is distributed in their layers. 
We do not attempt to achieve the best metaphor detection results but to analyze layers of PLMs to test if they contain the necessary information to perform this task.
In trying to answer this question,  we apply probing methods, discussed as follows, to focus on the representation itself by freezing the PLM parameters and training classifiers on top.

We hypothesize that metaphorical information does exist in PLM layers and more in the middle layers.
As we discussed earlier, metaphor detection depends on contrast prediction between source and target domains of a token. We assume that this prediction is made mainly based on the initial layers of PLM representations of either the token itself or its context or both. In higher layers of PLMs, the representations are dominated by contextual information, making it hard to retrieve the source domain, and so, reasoning about the contrast of the source and target domains becomes difficult.

\paragraph{Methods}
\label{sec:probing_methods}
We employ edge probing \cite{tenney2018you} and MDL \cite{voita-titov-2020-information}. Edge probing consists of a classifier in which word representations obtained from PLMs are fed as inputs
after projecting to 256-dimensional vectors first. 
The quality of the classifier illustrates how well the representations encode a specific linguistic knowledge. This method is designed for span-level tasks, i.e., the classifier can only access the representations of a limited part of the input sentence specified in the dataset. Edge probing has two pooler sections for making fixed-sized vectors; one pools representations across the words in the span and the other pools representations across the layers.

The Minimum Description Length (MDL) probing is based on information theory and combines the quality of the classifier and the amount of effort needed to achieve this quality. 

\citet{voita-titov-2020-information} propose two methods for computing MDL: ``\emph{variational coding}" and ``\emph{online coding.}" 
The former computes the complexity of the classifier with a Bayesian model. In the latter, the classifier is trained gradually on different portions of the dataset, and the code length will be the sum of the cross-entropies, each for a data portion. \citet{voita-titov-2020-information} show that the two methods' results are consistent with each other. 
Accordingly, we opted for the ``\emph{online coding}" method since it is more straightforward in implementation.
Since the code length is related to the size of the dataset $N$, we report the ``\emph{compression}", which is equal to 1 for a random classifier and larger for better models, and is defined as: $compression = \frac{N\cdot\log_2(K)}{\textsc{MDL}}$
See extra details in \citet{voita-titov-2020-information}.

\subsection{Generalization}
\label{sec:generalization}

To see if PLMs encode generalizable metaphorical knowledge, we evaluate them in settings where testing and training data are in different distributions.
We explore transferability analysis across languages and datasets as two sources of distribution.
We explain each in the following sections.

\begin{table*}[h!t!]
\centering
\scalebox{0.98}{\begin{tabular}{ll}
\toprule
VUA Verbs & 
\makecell[cl]{
    He \textbf{[finds]$_1$} it hard to communicate with people , not least his separated parents .$\rightarrow$ 1
    \\
    He finds it hard to \textbf{[communicate]$_1$} with people  , not least his separated parents . $\rightarrow$ 0
}
\\
\cline{2-2}
VUA POS & 
\makecell[cl]{
    They picked up power from a \textbf{[spider]$_1$} 's web of unsightly overhead wires . $\rightarrow$ 1
    \\
    They picked up power from a spider 's web of unsightly overhead \textbf{[wires]$_1$} . $\rightarrow$ 0  
}
\\
\cline{2-2}
TroFi & 
\makecell[cl]{
    `` Locals \textbf{[absorbed]$_1$} a lot of losses , '' said Mr. Sandor of Drexel $\rightarrow$ nonliteral
    \\
    Vitamins could be passed right out of the body without being \textbf{[absorbed]$_1$} $\rightarrow$ literal
}
\\
\cline{2-2}
LCC & 
\makecell[cl]{
    Lawful gun ownership is not a \textbf{[disease]$_1$} . $\rightarrow$ 3.0 
    \\
    But the Supreme Court says it's not a way to \textbf{[hurt]$_1$} the Second Amendment $\rightarrow$ 2.0 
    \\
    Is he angry that gun rights \textbf{[progress]$_1$} has been done without him? $\rightarrow$ 1.0 
    \\
    I mean the 2nd amendment \textbf{[suggests]$_1$} a level playing field for all of us. $\rightarrow$ 0.0
}
\\
\bottomrule
\end{tabular}
}
\caption{Examples of sentences, spans, and target labels for each probing dataset.}
\label{tab:metaphor_datasets_examples}
\end{table*}

\subsubsection{Cross-lingual}
\label{sec:cross_lingual}
Multilingual encoders project the representations in multiple languages into a shared space so that semantically similar words and sentences across languages end up close to each other. 
If we use a multilingual PLM model,
and our classifier shows that representations in language $S$ are informative about metaphoricity, what happens if we apply this classifier to the representations in language $T$?
We hypothesize that if the representation is rich in both languages, the annotation of metaphor is consistent, and the concept of metaphor is transferable across languages, then the classifier would be able to predict metaphoricity in language $T$ from what it learns in $S$.

When testing cross-lingual generalization, the linguistic and cultural differences of metaphoricity are important as well. 
We assume that metaphors are conceptualized in a similar process across languages, and metaphor detection is defined consistently.
The lexicalization is, of course, different, but that is something that multilingual PLMs are supposed to handle to some extent.

\begin{table}[t]
\centering
\scalebox{0.98}{\begin{tabular}{llr}
\toprule
dataset & POS & \multicolumn{1}{c}{Sizes} \\
\midrule
LCC (en)    & ALL & 28,096 / 4,014 / 8,028  \\
LCC (fa)    & ALL & 12,238 / 1,802 / 3,604  \\
LCC (es)    & ALL & 12,238 / 2,236 / 4,474  \\
LCC (ru)    & ALL & 12,238 / 1,748 / 3,498 \\
TroFi       & V & 3,838  / \hspace{0.75em}548   / 1,096 \\
VUA Verbs   & V & 9,176  / 1,310 / 2,622\\
VUA POS     & ALL & 21,036 / 3,006 / 6,010 \\
\bottomrule
\end{tabular}
}
\vspace{-1ex}
\caption{Statistics of the datasets. We label-balance each to have 50\% metaphor. Number of instances for train~/~dev~/~test sets and the types of POS are given as well. N: Noun, V: Verb,
ALL: Noun, Verb, Adjective, Adverb.
}
\label{tab:dataset_stat}
\end{table}
\begin{table*}[h]
\begin{center}
\begin{tabular}{ l | c c | c c | c c | c c } 
 \toprule
    & \multicolumn{2}{c|}{\textbf{Baseline}} & \multicolumn{2}{c|}{\textbf{BERT}} & \multicolumn{2}{c|}{\textbf{RoBERTa}} & \multicolumn{2}{c}{\textbf{ELECTRA}} \\
    \textbf{Dataset} & Acc. & Comp. & Acc. & Comp. & Acc. & Comp. & Acc. & Comp. \\
     \midrule
     LCC (en)   &  74.86  & 1.05\textsubscript{2} &  88.25       & 1.85\textsubscript{6} & 88.06 & 1.96\textsubscript{5}      & \bf{89.30} & \bf{2.05\textsubscript{5}}\\
     TroFi      &  67.34  & 1.01\textsubscript{4} &  \bf{68.58}  & 1.07\textsubscript{4} & 68.46 & \bf{1.09\textsubscript{6}} & 68.07      & 1.08\textsubscript{3}  \\
     VUA POS    &  65.92  & 1.03\textsubscript{0} &  80.32       & 1.43\textsubscript{5} & 81.72 & 1.48\textsubscript{6}      & \bf{83.03} & \bf{1.51\textsubscript{4}} \\ 
     VUA Verbs  &  65.97  & 1.04\textsubscript{9} &  78.29       & 1.28\textsubscript{9} & 78.88 & \bf{1.34\textsubscript{5}} & \bf{79.96} & 1.31\textsubscript{4}  \\ 
 \bottomrule
\end{tabular}
\end{center}
\caption{
Edge probing accuracy results for various metaphoricity datasets in BERT, RoBERTa, and ELECTRA. Baseline is a randomly initialized BERT. The edge probing results are the average of three runs. The compression result is the best across layers, and the subscript denotes the best layer.
}

\label{fig:met_det_f1}
\end{table*}

\subsubsection{Cross-dataset}
When training and testing on the same distribution, any learning model often uses heuristics and annotation biases. 
The consequence is the recurring overestimation of the capabilities of PLMs in doing hard tasks. This might be the case for our probing experiments as well.
Therefore, another generalization dimension we consider is cross-dataset transfer, i.e., training on dataset $S$ and testing on dataset $T$. $S$ and $T$ could be annotated by different people with possibly different goals in mind, and their raw sentences could come from different domains.
However, they must be annotated for the same task of metaphor detection.

In our case, the four datasets discussed more in \S \ref{sec:datasets} 
differ in their distribution of the candidate POS types (e.g., TroFi is only verbs, but LCC is not).
Further, the annotation process is different as each follows its own guidelines. However, the essential task of metaphor detection, i.e., distinguishing metaphor and literal usages, is the same for all.
Therefore, we expect some transferability across datasets but with differences aligned with their mismatches.

\section{Experimental Setup and Results}

\subsection{Datasets and Setup}
\label{sec:datasets}
\paragraph{Datasets} 
We use four metaphor detection datasets in our study.
The annotations of LCC \cite{mohler2016lcc} are done mostly on web crawled data as well as news corpora.
It provides metaphoricity scores including 0 as no , 2 as conventional, and 3 as clear metaphor.\footnote{1 is weak metaphor and as \citet{mohler2016lcc} describe metaphors with $0.5\leq score < 1.5$ as unclear, we ignore it.}
We use the examples with score 0 as literal, and others as metaphor.

TroFi dataset \cite{birke-sarkar-2006-clustering} consists of metaphoric and literal usages of 51 English verbs from WSJ.
VUA \cite{steen2010method} corpus consists of words in the academic, fiction, and news subdomains of the British National Corpus (BNC).
The authors published two versions: VUA POS and VUA Verbs.

LCC contains annotations in four languages: English, Russian, Spanish, and Farsi.
The other three datasets, TroFi, VUA Verbs and VUA POS, are in English only.
We have label-balanced all the datasets to get a more straightforward interpretation of results (the accuracy of a fair-coin random baseline is 50\% in all cases) and have split the datasets to train / dev / test sets with ratios of 0.7 / 0.1 / 0.2.

The statistics of the datasets are shown in Table \ref{tab:dataset_stat}.
Example sentences with the corresponding annotations can be seen in Table~\ref{tab:metaphor_datasets_examples}.

\paragraph{Setup}
In implementing the edge probe, we use batch size = 32 and learning rate = 5e-5 and train for five epochs in all experiments.
For the MDL probe, the same structure of edge probing is employed.
We apply a logarithm to the base two instead of the natural logarithm in cross-entropy loss to have all the obtained code lengths in bits (see extra details in \citealt{voita-titov-2020-information}). Our experiments are done using the GPUs provided by Google Colab free and pro.

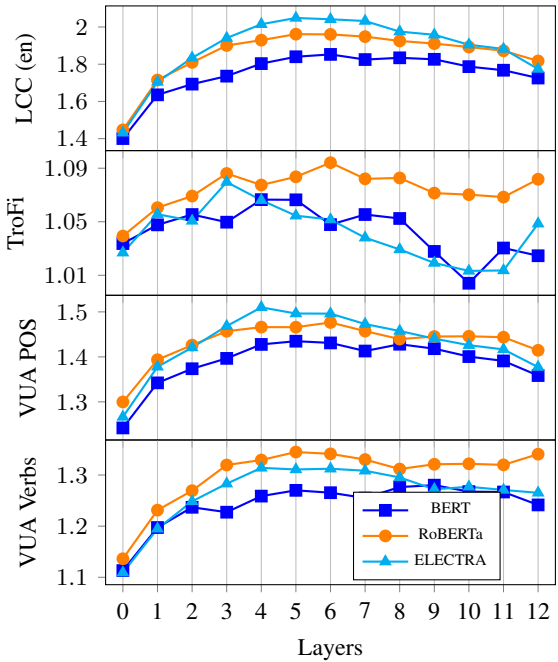
\begin{figure}[t]
\centering
    
    \begin{tikzpicture}
    \begin{groupplot}[
        group style={
            group name=my plots,
            group size=1 by 4,
            xlabels at=edge bottom,
            xticklabels at=edge bottom,
            vertical sep=0pt,
        },
        footnotesize,
        width=7.5cm,
        height=3.5cm,
        xlabel=Layers,
        xmin=-0.5, xmax=12.5,
        scaled y ticks=false,
        yticklabel style={/pgf/number format/fixed},
        legend style={font=\tiny, at={(0.55,0.03)}, anchor=south west},
        xtick={0,1,...,12},
        xticklabels={0,1,...,12},
        tickpos=left,
        ytick align=outside,
        xtick align=outside,
        xmajorgrids,
    ]

    \nextgroupplot[ylabel={LCC (en)}]
    \addplot[color=blue, mark=square*, thick]
        coordinates {
        (0, 1.4005438774758792)(1, 1.6354757027961522)(2, 1.6930593228028452)(3, 1.735948034965658)(4, 1.8033186795645317)(5, 1.8401051875231385)(6, 1.8527167376555365)(7, 1.825058272554039)(8, 1.8344422865917869)(9, 1.8264104377328925)(10, 1.7870196174060824)(11, 1.7677261068542898)(12, 1.725759134433315)
        };
    \addplot[color=orange, mark=*, thick]
        coordinates {
        (0, 1.447321549918507)(1, 1.7151397907959427)(2, 1.8095432057476661)(3, 1.9005985747034442)(4, 1.9292852356249413)(5, 1.9615513457451994)(6, 1.9602943547436043)(7, 1.9482657796203136)(8, 1.925173067447919)(9, 1.9106710979658557)(10, 1.8923532645745877)(11, 1.8723090361087085)(12, 1.8176165623955305)
        };
    \addplot[color=cyan, mark=triangle*, thick]
        coordinates {
        (0, 1.4317707469225562)(1, 1.7047181339658026)(2, 1.8342330902320956)(3, 1.9407207306955334)(4, 2.0148446112932703)(5, 2.048465344856003)(6, 2.041508822895552)(7, 2.0320363805606267)(8, 1.9752590203065998)(9, 1.9580609893257965)(10, 1.9048134350659751)(11, 1.8819450893329732)(12, 1.7742498009040077)
        };
        
    \nextgroupplot[ylabel={TroFi}, ytick={1.01, 1.05, 1.09}]
    \addplot[color=blue, mark=square*, thick]
        coordinates {
        (0, 1.0336987709778616)(1, 1.0476745094341262)(2, 1.0553858627168418)(3, 1.0496687337290014)(4, 1.066534745209823)(5, 1.066375597489138)(6, 1.0478583119503626)(7, 1.0553829359038764)(8, 1.0524328945214594)(9, 1.0278260098471723)(10, 1.0040459801792894)(11, 1.0303671317773775)(12, 1.0245932221968699)
        };
    \addplot[color=orange, mark=*, thick]
        coordinates {
        (0, 1.0393294215602051)(1, 1.0605132614496688)(2, 1.0691887001435245)(3, 1.0860757503576406)(4, 1.0774502051979202)(5, 1.0835584751494194)(6, 1.0941327069196611)(7, 1.0820521552888365)(8, 1.0827003892409532)(9, 1.0713938003030823)(10, 1.0702305116488573)(11, 1.068429582489663)(12, 1.0817586851395207)
        };
    \addplot[color=cyan, mark=triangle*, thick]
        coordinates {
        (0, 1.026868206159636)(1, 1.0556202056195472)(2, 1.0506406644649098)(3, 1.079639574006288)(4, 1.0660752300869847)(5, 1.0545162594867055)(6, 1.0516366984251675)(7, 1.038057260681288)(8, 1.029316842076766)(9, 1.0191791422354726)(10, 1.0132710362183273)(11, 1.0136071543911231)(12, 1.048543196650272)
        };
        
    \nextgroupplot[ylabel={VUA POS}]
    \addplot[color=blue, mark=square*, thick]
        coordinates {
        (0, 1.2420251349398113)(1, 1.3422127592821353)(2, 1.373624166924647)(3, 1.3966393304900004)(4, 1.427675493139513)(5, 1.434851145495114)(6, 1.4310173540740556)(7, 1.412986657952114)(8, 1.4281247622853104)(9, 1.4182166505587164)(10, 1.4006401032726772)(11, 1.3910141424544784)(12, 1.3584418326667531)
        };
    \addplot[color=orange, mark=*, thick]
        coordinates {
        (0, 1.2998017704054097)(1, 1.3936469782502547)(2, 1.4261287114756012)(3, 1.4569672531110875)(4, 1.4661448090016187)(5, 1.4659963891670458)(6, 1.4764762165332335)(7, 1.4572672676642433)(8, 1.4394911003574002)(9, 1.4452341953385874)(10, 1.445723666937929)(11, 1.443727198489866)(12, 1.4145776162008168)
        };
    \addplot[color=cyan, mark=triangle*, thick]
        coordinates {
        (0, 1.266498624875309)(1, 1.3778788034073814)(2, 1.42083285735495)(3, 1.4683958245800606)(4, 1.5100599990215162)(5, 1.496397314107032)(6, 1.4960921746926406)(7, 1.472974228905899)(8, 1.457503387697839)(9, 1.4405383207638656)(10, 1.4259624812439262)(11, 1.416572893718255)(12, 1.377258026043193)
        };
    
    \nextgroupplot[ylabel={VUA Verbs}, ytick={1.1, 1.2, 1.3}]
    \addplot[color=blue, mark=square*, thick]
        coordinates {
        (0, 1.1131760679308216)(1, 1.1973330911842532)(2, 1.2368947299193973)(3, 1.2272994017543324)(4, 1.2587804430184024)(5, 1.2700418686649377)(6, 1.265415722546408)(7, 1.2549782328122956)(8, 1.276486824453816)(9, 1.2801868099473772)(10, 1.2675713226558913)(11, 1.2668018582648435)(12, 1.2415172566740509)
        };
    \addplot[color=orange, mark=*, thick]
        coordinates {
        (0, 1.1362858073844964)(1, 1.2313481524308358)(2, 1.2694083530875664)(3, 1.3195784933820969)(4, 1.329361459229296)(5, 1.3449280209346852)(6, 1.3414152818412846)(7, 1.3302309892367512)(8, 1.3117268824291335)(9, 1.3210409753713985)(10, 1.321947451631722)(11, 1.3198525081835013)(12, 1.340804370086055)
        };
    \addplot[color=cyan, mark=triangle*, thick]
        coordinates {
        (0, 1.109107836007414)(1, 1.194959381927426)(2, 1.2484957320710997)(3, 1.2830337073254774)(4, 1.314006951430418)(5, 1.310889568307815)(6, 1.3123018461205767)(7, 1.3083524124817862)(8, 1.2952741334062539)(9, 1.2720944197314865)(10, 1.2770474345834444)(11, 1.2708156549259033)(12, 1.2651697252150373)
        };

    \legend{BERT,RoBERTa,ELECTRA}
    
    \end{groupplot}
    \end{tikzpicture}

    \caption{MDL compression across layers of three PLMs in four metaphor detection datasets. Higher number means better quality and extractability.}
    \label{fig:mdl_probing_layers}
\end{figure}

\subsection{Probing Results}
Here, BERT \cite{devlin-etal-2019-bert}, RoBERTa \cite{liu2019roberta}, and ELECTRA \cite{clark2020electra} represent our PLMs. Due to our resource limitations, we conduct all experiments on the \emph{base} version of the models (12 layers, 768 hidden size, 110M parameters) implemented in HuggingFace's Transfomers \citep{wolf-etal-2020-transformers}.
We employ edge probing for evaluating overall metaphorical knowledge in our selected PLMs, and MDL for the layer-wise comparisons.
MDL is shown to be more effective for layer-wise probing \cite{fayyaz2021models}.

Table~\ref{fig:met_det_f1} shows the edge probing accuracy and MDL probing compression results for our three PLMs. Accordingly, RoBERTa and ELECTRA are shown to encode metaphorical knowledge better than BERT on both metrics.
This is consistent with their better performance on various tasks, acquired by having better pre-training objectives and / or enjoying more extensive pre-training data.
The higher probing quality of ELECTRA's representations, is also consistent with \citet{fayyaz2021models} results on various linguistic knowledge tasks, including dependency  labeling,  named  entity recognition, semantic role labeling, and coreference resolution.

MDL probing compression across layers is demonstrated in Figure \ref{fig:mdl_probing_layers}.
We see the numbers increase mostly at the first 3 to 6 layers, depending on the dataset, but it decreases afterwards\footnote{For RoBERTa and in the case of TroFi and VUA Verbs, we see exceptional increases in the last layers.}. In other words, metaphorical information is more concentrated in the middle layers, where the representations are relatively contextualized but not as much as higher layers.
To put this in perspective, we can consider \citet{tenney-etal-2019-bert} and \citet{fayyaz2021models} where the best layers for various linguistic knowledge tasks in BERT are within 4 and 9. This shows that metaphor detection in PLM representations can be resolved earlier than some basic linguistic tasks.

In \S\ref{sec:probing_intro}, we elaborated a hypothesis that the process of detecting metaphors is not very deep since what it needs to do is mainly  contrast prediction between source and target domains, and the deep layers do not represent the source domain well. Our reported probing results confirm that metaphor detection is not deep in PLM layers.
To further evaluate our reasoning, we probe the domain knowledge in PLM representations across layers. 
We employ LCC's annotation of source and target domains, and run a similar MDL probing on different PLMs but for domain prediction. The obtained results, shown in Figure
\ref{fig:source_target_mdl_probing_layers} in appendix, demonstrate that the source domain information is represented in the initial layers (2-6), confirming that
the source domain is dominated by other information in higher layers.
On the other hand, target domain information generally increases across layers. Therefore, the middle layers can be the best place for contrasting source and target domains.

\begin{table*}[h]
\centering
\renewcommand{\arraystretch}{1.18}
\begin{tabular}{ l c c c c c c}
  \toprule
  & & \multicolumn{4}{c} {\textbf{Train Lang}}\\
  & \multicolumn{1}{c} {} &
  \multicolumn{1}{c} {en} &
  \multicolumn{1}{c} {es} &
  \multicolumn{1}{c} {fa} &
  \multicolumn{1}{c} {ru} \\
  \hline
  \multirow{4}{*}{\begin{sideways}\textbf{Test Lang}\end{sideways}} 
  & en & \textbf{85.14} (65.37)      & 79.31 (52.71)            & 77.59 (50.22)          & \underline{80.51} (52.40)\\
  & es & 79.40 (53.17)                & \textbf{84.59} (66.09)  & 76.70 (50.32)          & \underline{79.68} (53.32)\\
  & fa & 75.70 (50.07)                & 75.29 (52.65)           & \textbf{81.04} (65.91) & \underline{77.14} (50.36)\\
  & ru & \underline{83.92} (53.25)   & 80.54 (51.48)            & 76.61 (51.05)          & \textbf{88.36} (67.98)\\
  \toprule
\end{tabular}
\vspace{-1ex}
\caption{Cross-lingual metaphor detection accuracies after five epochs of training for XLM-R and (its random version).
For each test language, we bold its in-distribution (e.g., en $\rightarrow$ en), and underline the best out-of-distribution (e.g., ru $\rightarrow$ en) numbers.
}
\label{tab:cross-lingual}
\end{table*}

\subsection{Generalization Experiments}

As our PLMs, we use XLM-R \cite{conneau-etal-2020-unsupervised} for cross-lingual and BERT for cross-dataset experiments.
To compare the cross-lingual and cross-dataset transferability, in \S \ref{sec:cross_compare}, we employ the same setup, including using XLM-R as PLM for both.
The results in \S \ref{sec:cross_lingual_res} and  \ref{sec:cross_dataset_res} are not comparable.
We apply the same edge probing architecture as in the probing experiments.
We sometimes refer to both language and dataset as \emph{distribution}.

We run two experiments for each case of a source distribution $S$ and a target distribution $T$: one with the PLM and one with a randomized version of the PLM where weights are set to random values. Randomly initialized Transformers with the same architecture as PLMs are common baselines in the community. The difference between the two gives evidence about the helpfulness of the encoded knowledge gained during pre-training in doing the task. When  $S = T$, this effect is measured for in-distribution and when $S \neq T$, for out-of-distribution generalization.
Comparing results of in-distribution (e.g., training and testing on English data) and out-of-distribution (e.g., training on Spanish and testing on English) setups demonstrates how
generalizable the metaphorical knowledge in PLM is and 
how consistent the annotations are.

\begin{table*}[h]
\centering
\renewcommand{\arraystretch}{1.18}
\begin{tabular}{ l l c c c c c}
  \toprule
  & & \multicolumn{4}{c} {\textbf{Train Dataset}}\\
  & \multicolumn{1}{c} {} &
  \multicolumn{1}{c} {LCC(en)} &
  \multicolumn{1}{c} {TroFi} &
  \multicolumn{1}{c} {VUA POS} &
  \multicolumn{1}{c} {VUA Verbs} \\
  \hline
  \multirow{4}{*}{\begin{sideways}\textbf{Test Dataset}\end{sideways}} 
  & LCC(en)   & \textbf{84.26} (54.93)       & 62.04 (50.05)          & 70.35 (50.69)               & \underline{70.37} (50.14) \\
  & TroFi     & 59.49 (50.58)                & \textbf{68.73} (64.96) & 55.38 (49.45)               & \underline{59.67} (53.68) \\
  & VUA POS   & 62.23 (51.47)                & 55.29 (50.47)          & \textbf{76.86} (56.01)      & \underline{71.6} (53.47) \\
  & VUA Verbs & 60.20 (50.88)                & 54.55 (51.73)          & \underline{72.6} (56.01)    & \textbf{75.21} (60.03) \\
  \toprule
\end{tabular}
\vspace{-1ex}
\caption{Cross-dataset edge probing accuracy results on BERT is shown in pairs: pre-trained model and, in the parenthesis, the randomly initialized model. 
We set the training size to the minimum among datasets, i.e., TroFi.
For each test dataset, we bold its in-distribution (e.g., VUA Verbs $\rightarrow$ VUA Verbs), and underline the best out-of-distribution (e.g., VUA POS $\rightarrow$ VUA Verbs) numbers.
}
\label{tab:cross-dataset}
\end{table*}
\subsubsection{Cross-lingual}
\label{sec:cross_lingual_res}
The four LCC datasets corresponding to four languages are used here.
We subsample from the datasets to have the same number of examples in the training sets, i.e., 12,238 which is the size of the Russian training set.
The results are shown in Table~\ref{tab:cross-lingual}.
The random baseline is acquired using a randomly initialized XLM-R.

We observe that XLM-R significantly outperforms the random, confirming that metaphorical knowledge learned during the pre-training is transferable across languages.
This considerable transferability can be attributed to the ability of XLM-R to build language-universal representations useful for metaphoricity transfer.
Moreover, the innate similarities of metaphors in distinct languages can contribute to higher transferability, despite the lexicalization differences. E.g., analogizing a concept to a tool (en) occurs the same way in other languages like instrumento (es), \FR{\smallابزار} (fa) and \foreignlanguage{russian}{инструмент} (ru).
Finally, the constraints of the dataset producers in, for instance, keeping the languages in relatively similar target and source domains, could be influential. (See Figures \ref{fig:source_concept} and \ref{fig:target_concept}).

An interesting observation is that training on Russian shows the best out-of-distribution results when testing on other languages.  
We analyze this further.
First, we observe that LCC(ru) has almost the closest target domain distribution to all other languages (See Table \ref{tab:cross-lingual_target_dist} in Appendix). 

Second, the reported results can also be influenced by the amount of data from each of these languages in the pre-training data of \nohyphens{XLM-R}. Russian has the second largest size after English \cite{conneau-etal-2020-unsupervised}.
Finally, for English, the higher-resource language with closer target domain distribution, we find out that there are considerable number of examples in the LCC(en) related to ``GUNS" and ``CONTROL\_OF\_GUNS".
These domains are not covered in other LCC datasets (See Figure \ref{fig:target_concept} in Appendix).

\subsubsection{Cross-dataset}
\label{sec:cross_dataset_res}
Similar to the cross-lingual evaluations, here we have four datasets used as sources and targets. We set the train size of each to the minimum of all, i.e., 3,838.
For each pair, we run two experiments: one with randomized and one with pre-trained BERT as our PLM.
Results are shown in Table \ref{tab:cross-dataset}.

PLM is much better than random in all out-of-distribution cases, suggesting the presence of generalizable metaphorical information.
As expected, VUA Verbs and POS achieve the best results when mutually tested, because, apart from the POS, they have the same distribution.
VUA datasets and LCC(en) show good transferability, but the gap with in-distribution results is still considerable ($>$13\% absolute). 
VUA Verbs is the best source for TroFi, likely because of the POS match between them.
Overall, apart from the two VUA datasets, the gap between in- and out-of-distribution performance is large. 

The random PLM accuracies range from about 54\%-64\% and 50\%-56\% for in- and out-of-distribution cases. We hypothesize that this drop in the out-of-distribution is related to the annotation biases, which a randomly initialized classifier can leverage better when testing and training sets are from the same distribution. 
When the sets have different distributions, the biases do not transfer well.

\subsubsection{Comparing cross-dataset and cross-lingual}
\label{sec:cross_compare}
\begin{table}[h]
\centering
\tabcolsep=0.11cm
\renewcommand{\arraystretch}{1.18}
\scalebox{0.95}{\begin{tabular}{ c c c c }
  \toprule
  \multicolumn{1}{c} {LCC(en)} &
  \multicolumn{1}{c} {LCC(es)} &
  \multicolumn{1}{c} {LCC(fa)} &
  \multicolumn{1}{c} {LCC(ru)} \\
  82.31 & 78.02  & 77.3 & 78.04 \\
\midrule
  \multicolumn{1}{c} {TroFi} &
  \multicolumn{1}{c} {VUA POS} &
  \multicolumn{1}{c} {VUA Verbs}\\
  60.54 & 68.61 & 67.15 & \\
  \bottomrule
\end{tabular}
}
\vspace{-1ex}
\caption{Comparing cross-dataset and cross-lingual scenarios using the same model (XLM-R), training size, testing set, i.e., LCC(en), and different training sources.
}
\label{tab:cross_ling_data_mix}
\end{table}

As additional transferability analysis, we compare cross-lingual and cross-dataset results,
by using XLM-R and evaluating different training sources on
LCC(en) test set.
We make the size of each train set to be the same (3,838).
The results are shown in Table \ref{tab:cross_ling_data_mix}, where the first and second rows belong to cross-lingual and cross-dataset, respectively.
To base our results, we include the in-distribution result of training on LCC(en), i.e., 82.31\%.

Clearly, there is a substantial gap between cross-lingual and cross-dataset accuracies.
The annotation guideline is consistent in the LCC language datasets, while for the cross-dataset settings, we have datasets that differ in many aspects, including annotation procedure and definitions, covered part-of-speeches (e.g., Trofi and VUA Verbs vs. LCC and VUA POS) and sentence lengths (LCC: 25.9, VUA: 19.4, Trofi: 28.3).

\section{Discussion and Conclusion }
Metaphors are important in human cognition, and if we seek to build cognitively inspired or plausible language understanding systems, we need to work more on their best integration in the future. Therefore, any work in this regard is impactful.

Our probing experiments showed that PLMs do in fact represent the information necessary to do the task of metaphor detection.
We assume this information is related to metaphorical knowledge learned during pre-training. Further, the layer-wise analysis confirmed our hypothesis that middle layers are more informative.

Even though our probing experiments did show that metaphorical knowledge is present in PLMs, it was still unclear if this knowledge is generalizable beyond the training data.
So, to probe the probe and evaluate generalization, we ran cross-lingual and cross-dataset experiments.
Our results showed that the transferability across languages works quite well for the four languages in LCC annotation.
However, when the definitions and annotations were inconsistent across different datasets, the cross-dataset results were not satisfactory. 

Overall, we conclude that metaphorical knowledge does exist in PLM representations and in middle layers mainly, and it is transferable if the annotation is consistent across training and testing data. 
We will explore more the cross-lingual transfer of metaphors and the impact of cross-cultural similarities in the future.
Also, the application of metaphorical knowledge for text generation is something important that we will address.

\section*{Acknowledgements}
We would like to thank the anonymous reviewers and action editors who helped us greatly in improving our work with their comments.

\bibliographystyle{acl_natbib}
\bibliography{anthology,acl2021}

\appendix
\counterwithin{figure}{section}
\counterwithin{table}{section}
\section{Appendices}
\label{sec:appendix}

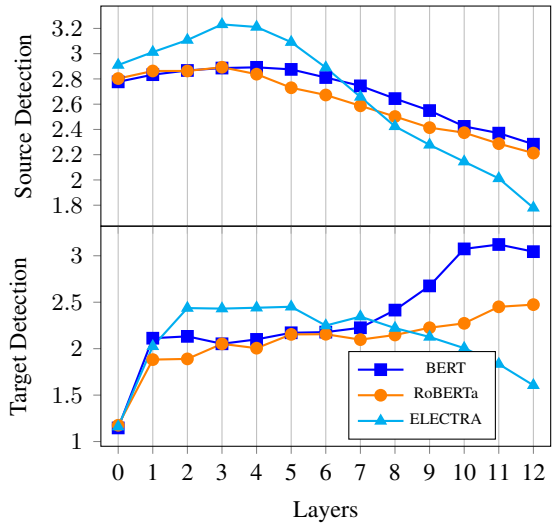
\begin{figure}[h]
\centering
    
    \begin{tikzpicture}
    \begin{groupplot}[
        group style={
            group name=my plots,
            group size=1 by 2,
            xlabels at=edge bottom,
            xticklabels at=edge bottom,
            vertical sep=0pt,
        },
        footnotesize,
        width=7.5cm,
        height=4.5cm,
        xlabel=Layers,
        xmin=-0.5, xmax=12.5,
        scaled y ticks=false,
        yticklabel style={/pgf/number format/fixed},
        legend style={font=\tiny, at={(0.55,0.03)}, anchor=south west},
        xtick={0,1,...,12},
        xticklabels={0,1,...,12},
        tickpos=left,
        ytick align=outside,
        xtick align=outside,
        xmajorgrids,
    ]

    \nextgroupplot[ylabel={Source Detection}]
    \addplot[color=blue, mark=square*, thick]
        coordinates {
        (0, 2.7767808730163304)(1, 2.833672047663791)(2, 2.866949425682184)(3, 2.886097503078181)(4, 2.891786523032616)(5, 2.875452363079488)(6, 2.8110062488630843)(7, 2.7449813292515146)(8, 2.6441674208614914)(9, 2.549355243760758)(10, 2.424173484012746)(11, 2.3706374024367074)(12, 2.283095144557525)
        };
    \addplot[color=orange, mark=*, thick]
        coordinates {
        (0, 2.8029397309743507)(1, 2.8628575983063596)(2, 2.863067933530763)(3, 2.8923929140039375)(4, 2.8375803668694224)(5, 2.7300684947522367)(6, 2.6727078803172013)(7, 2.5870348601113395)(8, 2.5022084625411787)(9, 2.4141404419002814)(10, 2.3743045657754736)(11, 2.287444790007937)(12, 2.212718907681295)
        };
    \addplot[color=cyan, mark=triangle*, thick]
        coordinates {
        (0, 2.909846433547726)(1, 3.012201086857984)(2, 3.1078401817616657)(3, 3.2317916143780905)(4, 3.2111703631068433)(5, 3.091173722614377)(6, 2.8896432671677603)(7, 2.657264081525527)(8, 2.4243019636621046)(9, 2.277827281486501)(10, 2.144859019698162)(11, 2.0121305205814615)(12, 1.7792215337785782)
        };
        
    Target
    \nextgroupplot[ylabel={Target Detection}]
    \addplot[color=blue, mark=square*, thick]
        coordinates {
        (0, 1.1466721661393489)(1, 2.112507566101032)(2, 2.1323648191336013)(3, 2.052442980233251)(4, 2.0997047446376023)(5, 2.171054539555421)(6, 2.178037430189697)(7, 2.224967528018063)(8, 2.4151844936028164)(9, 2.675536091964247)(10, 3.073886968399618)(11, 3.1211765171387986)(12, 3.04426565833024)
        };
    \addplot[color=orange, mark=*, thick]
        coordinates {
        (0, 1.1753852363948811)(1, 1.882160308968557)(2, 1.889779643675263)(3, 2.051667329840013)(4, 2.0050921532690658)(5, 2.153557806604654)(6, 2.155142249495772)(7, 2.09719688520241)(8, 2.147134830553604)(9, 2.225245221452384)(10, 2.2725651275183805)(11, 2.450451635242545)(12, 2.4738808303948456)
        };
    \addplot[color=cyan, mark=triangle*, thick]
        coordinates {
        (0, 1.1604099347755226)(1, 2.0249542022705502)(2, 2.43686359552071)(3, 2.431039445972411)(4, 2.4405011030126724)(5, 2.450939294709398)(6, 2.248666700157353)(7, 2.344872429552844)(8, 2.221523518134226)(9, 2.1259549067845493)(10, 2.0054265825762903)(11, 1.8339614988373756)(12, 1.6057778656233805)
        };

    \legend{BERT,RoBERTa,ELECTRA}
    
    \end{groupplot}
    \end{tikzpicture}

    \caption{MDL probing compression across layers for source and target domain detection for LCC(en) dataset.}
    \label{fig:source_target_mdl_probing_layers}
\end{figure}
\begin{table}[h]
\centering
\begin{tabular}{ c | c c c c}
    \toprule
    \multicolumn{1}{c} {} &
    \multicolumn{1}{c} {en} &
    \multicolumn{1}{c} {es} &
    \multicolumn{1}{c} {fa} &
    \multicolumn{1}{c} {ru} \\
    \hline
    en & 0.0000           &        &                    &        \\
    es & \textbf{0.1622}  & 0.0000 &                    &        \\
    fa & 0.1851           & 0.1688 & 0.0000             &        \\
    ru & 0.1833           & 0.2239 & \underline{0.2244} & 0.0000 \\
    \bottomrule
\end{tabular}
\vspace{-1ex}
\caption{Jensen–Shannon divergence between source domain frequency distribution of different languages. The datasets are the same ones used in cross-lingual experiments where train set sizes are set to 12,238. Bold denotes the closest distributions and underline denotes the furthest distributions.}
\label{tab:cross-lingual_src_dist}
\end{table}
\begin{table}[h]
\centering
\begin{tabular}{ c | c c c c}
    \toprule
    \multicolumn{1}{c} {} &
    \multicolumn{1}{c} {en} &
    \multicolumn{1}{c} {es} &
    \multicolumn{1}{c} {fa} &
    \multicolumn{1}{c} {ru} \\
    \hline
    en & 0.0000             &                 &                    &        \\
    es & 0.4116             & 0.0000 &                 &        \\
    fa & \underline{0.5004} & 0.2148          & 0.0000             &        \\
    ru & 0.4291             & \textbf{0.1209} & 0.2141 & 0.0000 \\
    \bottomrule
\end{tabular}
\vspace{-1ex}
\caption{Jensen–Shannon divergence between target domain frequency distribution of different languages. The datasets are the same ones used in cross-lingual experiments where train set sizes are set to 12,238. Bold denotes the closest distributions and underline denotes the furthest distributions.}
\label{tab:cross-lingual_target_dist}
\end{table}
\begin{table*}[htbp]
\centering
\begin{tabular}{ccc}
Language & Sentence & Annotations \\
\toprule
fa & 
\makecell[cr]{
    \FR{\small
    اما امريكا در افغانستان، از همان آغاز، با 
    }
    \\
    \FR{\small
    [
    سلاح
    ]$_1$
    [
    دموكراسي
    ]$_2$
    آمده است
    .
    }
    \\
}
&
\makecell[cl]{
    Score: 3.0 \\ 
    Src Concept: WAR(3.0)\\ 
    Target Concept: DEMOCRACY \\ 
    Polarity: NEUTRAL \\ 
    Intensity: 1.0
}
\\
\midrule
es & 
\makecell[cl]{
    [atorado]$_1$ en la [deuda]$_2$ pública
    \\
    y sin avances en Estado de Derecho
    \\
}
&
\makecell[cl]{
    Score: 3.0 \\ 
    Src Concept: BARRIER(3.0) \\ 
    Target Concept: DEBT \\ 
    Polarity: NEGATIVE \\ 
    Intensity: 2.0
}
\\
\midrule
ru & 
\makecell[cl]{
    \foreignlanguage{russian}{
    Мировые [деньги]$_2$ [мечутся]$_1$ , 
    }
    \\
    \foreignlanguage{russian}{
    не зная , куда вложиться .
    }
    \\
}
&
\makecell[cl]{
    Score: 3.0 \\ 
    Src Concept: MOVEMENT(3.0) \\ 
    Target Concept: MONEY \\ 
    Polarity: NEGATIVE \\ 
    Intensity: 2.0
}
\\
\bottomrule
\end{tabular}
\caption{Examples of sentences, spans, and annotations for LCC dataset in Farsi, Spanish, and Russian.}
\label{tab:lcc_examples}
\end{table*}
\newpage
\clearpage
\begin{figure}[t!]
\centering
    \includegraphics[width=0.41\textwidth]{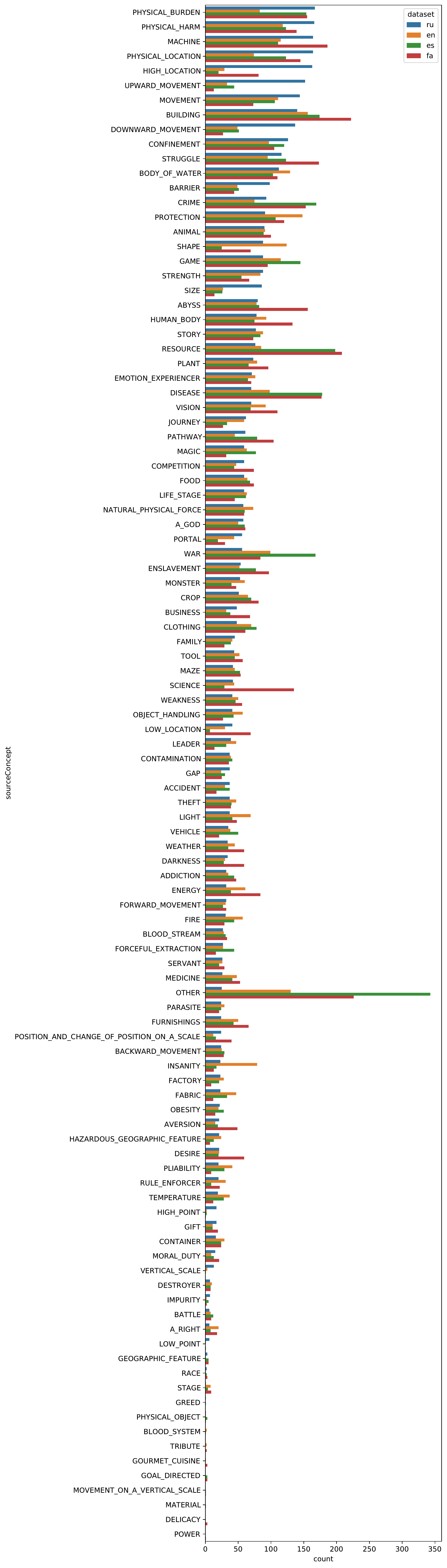}
    \caption{Source domain frequency in training set of cross-lingual datasets.}
    \label{fig:source_concept}
\end{figure}
\begin{figure}[t!]
\centering
    \includegraphics[width=0.476\textwidth]{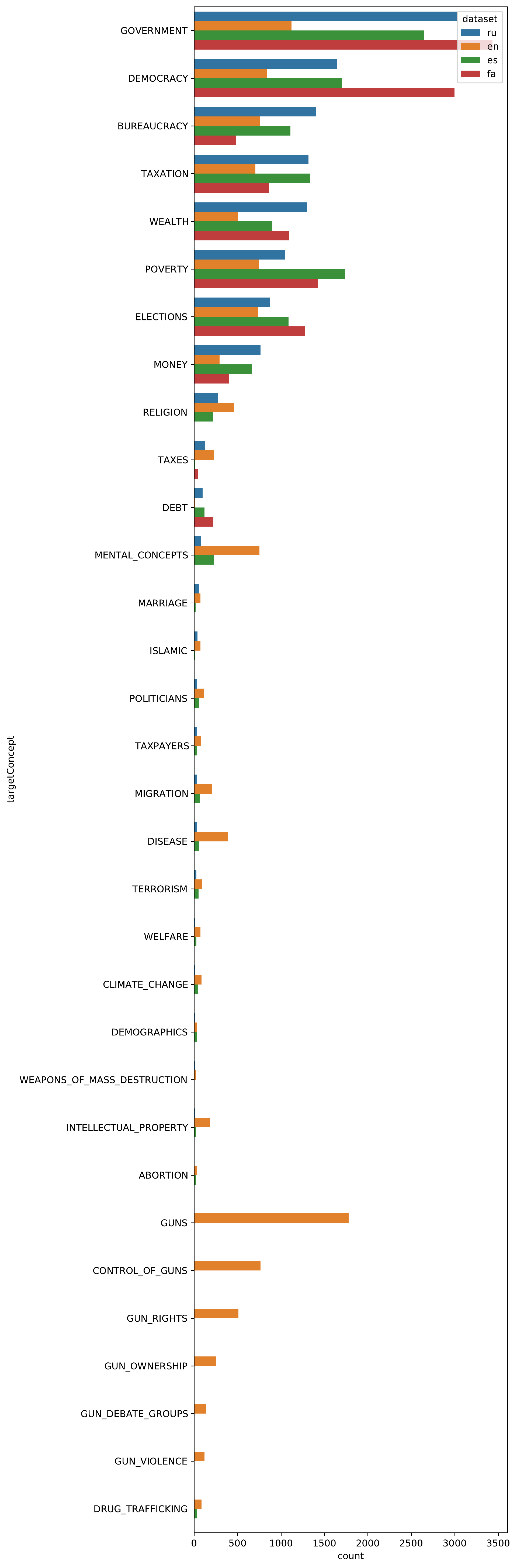}
    \caption{Target domain frequency in training set of cross-lingual datasets.}
    \label{fig:target_concept}
\end{figure}

\end{document}